\documentclass[letterpaper]{article}
\usepackage{iccc}

\usepackage{times}
\usepackage{helvet}
\usepackage{courier}
\usepackage[colorinlistoftodos]{todonotes}
\usepackage{algorithm}
\usepackage{comment}
\floatname{algorithm}{Example}
\title{DeepTingle}
\author{Ahmed Khalifa, Gabriella A. B. Barros \and Julian Togelius\\
Department of Computer Science and Engineering\\
New York University\\
Brooklyn, NY 11201 USA\\
ahmed.khalifa@nyu.edu, gabriella.barros@gmail.com \and  julian@togelius.com\\
}
\begin{document} 
\maketitle
\begin{abstract}
\begin{quote}
DeepTingle is a text prediction and classification system trained on the collected works of the renowned fantastic gay erotica author Chuck Tingle. Whereas the writing assistance tools you use everyday (in the form of predictive text, translation, grammar checking and so on) are trained on generic, purportedly ``neutral'' datasets, DeepTingle is trained on a very specific, internally consistent but externally arguably eccentric dataset. This allows us to foreground and confront the norms embedded in data-driven creativity and productivity assistance tools. As such tools effectively function as extensions of our cognition into technology, it is important to identify the norms they embed within themselves and, by extension, us. DeepTingle is realized as a web application based on LSTM networks and the GloVe word embedding, implemented in JavaScript with Keras-JS.
\end{quote}
\end{abstract}

\section{Introduction}




We live continuously computationally assisted lives. Computational assistance tools extend and scaffold our cognition through the computational devices, such as phones and laptops, that many of us keep close at all times. A trivial-seeming but important example is predictive text entry, also popularly known as autocomplete. The absence of regular keyboards on mobile devices have necessitated software which maps button-presses (or swipes) to correct words, and thus guesses what word we meant to write. In many cases, e.g. on the iPhone, the software also guesses what word you plan to write next and gives you the chance to accept the software's suggestion instead of typing the word yourself. Even when writing on a computer with a real keyboard, spell-checking software is typically running in the background to check and correct the spelling and sometimes the grammar of the text. In the structured domain of programming, Integrated Development Environments such as Eclipse or Visual Studio suggest what methods you want to call based on data-driven educated guesses. Relatedly, when shopping or consuming music or videos online, recommender systems are there to provide us with ideas for what to buy, watch or listen to next.

Beyond the relatively mundane tasks discussed above, there is a research vision of computational assistance with more creative tasks. The promise of computational creativity assistance tools is to help human beings, both professional designers and more casual users, to exercise their creativity better. An effective creativity assistance tool helps its users be creative by, for example, providing domain knowledge, assisting with computational tasks such as pattern matching, providing suggestions, or helping enforce constraints; and many other creativity assistance mechanisms are possible. This vision is highly appealing for those who want to see computing in the service of humanity. In the academic research community, creativity assistance tools are explored for such diverse domains as music~\cite{hoover2011interactively}, game levels~\cite{liapis2013sentient,smith2011tanagra,shaker2013ropossum}, stories~\cite{roemmele2015creative}, drawings~\cite{zhang2015drawcompileevolve}, and even ideas~\cite{llano2014baseline}.

There's no denying that many of these systems can provide real benefits to us, such as faster text entry, useful suggestion for new music to listen to, or the correct spelling for Massachusetts. However, they can also constrain us. Many of us have experienced trying to write an uncommon word, a neologism, or a profanity on a mobile device just to have it ``corrected'' to a more common or acceptable word. Word's grammar-checker will underline in aggressive red grammatical constructions that are used by Nobel prize-winning authors and are completely readable if you actually read the text instead of just scanning it. These algorithms are all too happy to shave off any text that offers the reader resistance and unpredictability. And the suggestions for new books to buy you get from Amazon are rarely the truly left-field ones---the basic principle of a recommender system is to recommend things that many others also liked.

What we experience is an algorithmic enforcement of norms. These norms are derived from the (usually massive) datasets the algorithms are trained on. In order to ensure that the data sets do not encode biases, ``neutral'' datasets are used, such as dictionaries and Wikipedia. (Some creativity support tools, such as Sentient Sketchbook~\cite{liapis2013sentient}, are not explicitly based on training on massive datasets, but the constraints and evaluation functions they encode are chosen so as to agree with ``standard'' content artifacts.) However, all datasets and models embody biases and norms. In the case of everyday predictive text systems, recommender systems and so on, the model embodies the biases and norms of the majority.

It is not always easy to see biases and norms when they are taken for granted and pervade your reality. Fortunately, for many of the computational assistance tools based on massive datasets there is a way to drastically highlight or foreground the biases in the dataset, namely to train the models on a completely different dataset. In this paper we explore the role of biases inherent in training data in predictive text 
algorithms through creating a system trained not on ``neutral'' text but on the works of Chuck Tingle.

Chuck Tingle is a renowned Hugo award nominated author of fantastic gay erotica. His work can be seen as erotica, science fiction, absurdist comedy, political satire, metaliterature, or preferably all these things and more at the same time. The books frequently feature gay sex with unicorns, dinosaurs, winged derrières, chocolate milk cowboys, and abstract entities such as Monday or the very story you are reading right now. The bizarre plotlines feature various landscapes, from paradise islands and secretive science labs, to underground clubs and luxury condos inside the protagonist's own posterior. The corpus of Chuck Tingle's collected works is a good choice to train our models on precisely because they so egregiously violate neutral text conventions, not only in terms of topics, but also narrative structure, word choice and good taste. They are also surprisingly consistent in style, despite the highly varied subjects. Finally, Chuck Tingle is a very prolific author, providing us with a large corpus to train our models on. In fact, the consistency and idiosyncracy of his literary style together with his marvelous productivity has led more than one observer to speculate about whether Chuck Tingle is actually a computer program, an irony not lost on us.

In this paper, we ask the question what would happen if our writing support systems did not assume that we wanted to write like normal people, but instead assumed that we wanted to write like Chuck Tingle. We train a deep neural net based on Long Short-Term Memory and word-level embeddings to predict Chuck Tingle's writings, and using this model we build a couple of tools (a predictive text system and a reimagining of literary classics) that assists you with getting your text exactly right, i.e. to write just like Chuck Tingle would have.

A secondary goal of the research is to investigate how well we can learn to generate text that mimics the style of Chuck Tingle from his collected works. The more general question is that of generative modeling of literary style using modern machine learning methods. The highly distinctive style of Tingle's writing presumably makes it easy to verify whether the generated text adheres to his style.

\section{Background}

This work builds on a set of methods from modern machine learning, in particular in the form of deep learning.

\subsection{Word Embedding}
Word embedding is a technique for converting words into a n-dimensional vector of real numbers, capable of capturing probabilistic features of the words in the current text. The primary goal is to reduce the dimensionality of the word space to a point where it can be easily processed. Each dimension in the vector represent a linguistic context, and the representation should preserve characteristics of the original word~\cite{goldberg2014word2vec}.

Such mappings have been achieved using various techniques, such as neural networks~\cite{bengio2003neural}, principal component analysis~\cite{lebret2013word}, and probabilistic models~\cite{globerson2007euclidean}. A popular method is skip-gram with negative-sampling training, a context-predictive approach implemented in word2vec models~\cite{mikolov2013distributed}. On the other hand, global vectors (GloVe) is a context-count word embedding technique~\cite{pennington2014glove}. GloVe captures the probability of a word appearing in a certain context in relation to the remaining text.

\subsection{Neural Networks and Recurrent Neural Networks}

Neural networks (NN) are a machine learning technique originally inspired by the way the human brain functions~\cite{hornik1989multilayer}. The basic unit of a NN is a neuron. Neurons receive vectors as inputs, and output values by applying a non linear function to the multiplication of said vectors and a set of weights. They are usually grouped in layers, and neurons in the same layer cannot be connected to each other. Neurons in a given layer are fully connected to all neurons in the following layer. NNs can be trained using the backpropagation algorithm. Backpropagation updates the network weights by taking small steps in the direction of 
minimizing the error measured by the network.

A recurrent neural network (RNN) is a special case of neural network. In a RNN, the output of each layer depends not only on the input to the layer, but also on the previous output. RNNs are trained using backpropagation through time (BPTT)~\cite{werbos1990backpropagation}, an algorithm that unfolds the recursive nature of the network for a given amount of steps, and applies a generic backpropagation to the unfolded RNN. Unfortunately, BPTT doesn't suit vanilla RNNs when they run for large amount of steps~\cite{hochreiter1998vanishing}. 
One solution for this problem
is the use of Long Short-Term Memory (LSTM). LSTMs were introduced by Sepp Hochreiter and J{\"u}rgen Schmidhuber (~\citeyear{hochreiter1997long}), and introduces a memory unit. The memory unit acts as a storage device for the previous input values. The input is added to the old memory state using gates. These gates control the percentage of new values contributing to the memory unit with respect to the old stored values. Using gates helps to sustain constant optimization through each time step.










\subsection{Natural Language Generation}

Natural language generation approaches can be divided into two categories: Rule- or template-based and machine learning ~\cite{tang2016context}. Rule-based (or template-based) approaches ~\cite{cheyer2014method,mirkovic2011dialogue} were considered norm for most systems, with rules/templates handmade. However, these tend to be too specialized, not generalizing well to different domains, and a large amount of templates is necessary to generate quality text even on a small domain. Some effort has been made towards generating the template based on a corpus, using statistical methods ~\cite{mairesse2010phrase,mairesse2014stochastic,oh2000stochastic}, but these still require a large amount of time and expertise.

Machine learning, in particular RNNs, has become an increasingly popular tool for text generation. Sequence generation by character prediction has been proposed using LSTM \cite{graves2013generating}) and multiplicative RNNs~\cite{sutskever2011generating}. Tang et al. (~\citeyear{tang2016context}) attempted associating RNNs and context-awareness in order to improve consistency, by encoding not only the text, but also the context in semantic representations. Context has also been applied in response generation in conversation systems ~\cite{sordoni2015neural,wen2015semantically}.

Similarly, machine learning is also used in machine translation ~\cite{sutskever2014sequence,cho2014properties,bahdanau2014neural}. These approaches tend to involve training a deep network, capable of encoding sequences of text from an original language in a fixed-length vector, and decoding output sequences to the targeted language.

\subsection{Creativity Assistance Tools}


Several works have been proposed to foster the collaboration between machine and user in creative tasks. Goel and Joyner argue that scientific discovery can be considered a creative task, and propose MILA-S, an interactive system with the goal of encouraging scientific modeling ~\cite{goel2015impact}. It makes possible the creation of conceptual models of ecosystems, which are evaluated with simulations.

CAHOOTS is a chat system capable of suggesting images as possible jokes ~\cite{wen2015omg}. STANDUP ~\cite{waller2009evaluating} assists children who use augmentative and alternative communication to generate puns and jokes.

Co-creativity systems can also help the creation of fictional ideas. Llano et al.(~\citeyear{llano2014baseline}) describe three baseline ideation methods using ConceptNet, ReVerb and bisociative discovery , while I-get ~\cite{ojha2015get} uses conceptual and perceptual similarity to suggest pairs of images, in order to stimulate the generation of ideas. 

DrawCompileEvolve ~\cite{zhang2015drawcompileevolve} is a mixed-initiative art tool, where the user can draw and group simple shapes, and make artistic choices such as symmetric versus assymetric. The system then uses uses neuroevolution to evolve a genetic representation of the drawing.

Sentient Sketchbook and Tanagra assist in the creation of game levels. Sentient Sketchbook uses user-made map sketches to generate levels, automate playability evaluations and provide various visualizations~\cite{liapis2013sentient,yannakakis2014mixed}. Tanagra uses the concept of rhythm to generate levels for a 2D platform~\cite{smith2010tanagra}.

Focusing on writing, we can highlight the Poetry Machine ~\cite{kantosalo2014isolation} and Creative Help~\cite{roemmele2015creative}. Both aim to provide suggestions to writers, assisting their writing process. The Poetry Machine creates draft poems based on a theme selected by the user. 
Creative Help uses case-based reasoning to search a large story corpus for possible suggestions ~\cite{roemmele2015creative}.









\section{DeepTingle}

This section discusses the methodology applied in DeepTingle. DeepTingle consists of two main components: the neural network responsible for the learning and prediction of words in the corpus, and a set of co-creativity tools aimed at assisting in the writing or style-transfer of text. The tools described (Predictive Tingle and Tingle Classics) are available online, at http://www.deeptingle.net.

Our training set includes all Chuck Tingle books released until November 2016: a total of 109 short stories and 2 novels (with 11 chapters each) to create a corpus of 3,044,178 characters. The text was preprocessed by eliminating all punctuation, except periods, commas, semicolons, question marks and apostrophes. The remaining punctuation marks, excluding apostrophes, were treated as separate words. Apostrophes were attached to the words they surround. For example, ``{I'm}'' is considered a single word.

\subsection{Network Architecture}

\begin{figure}[!b]
\centering
\includegraphics[height=3cm]{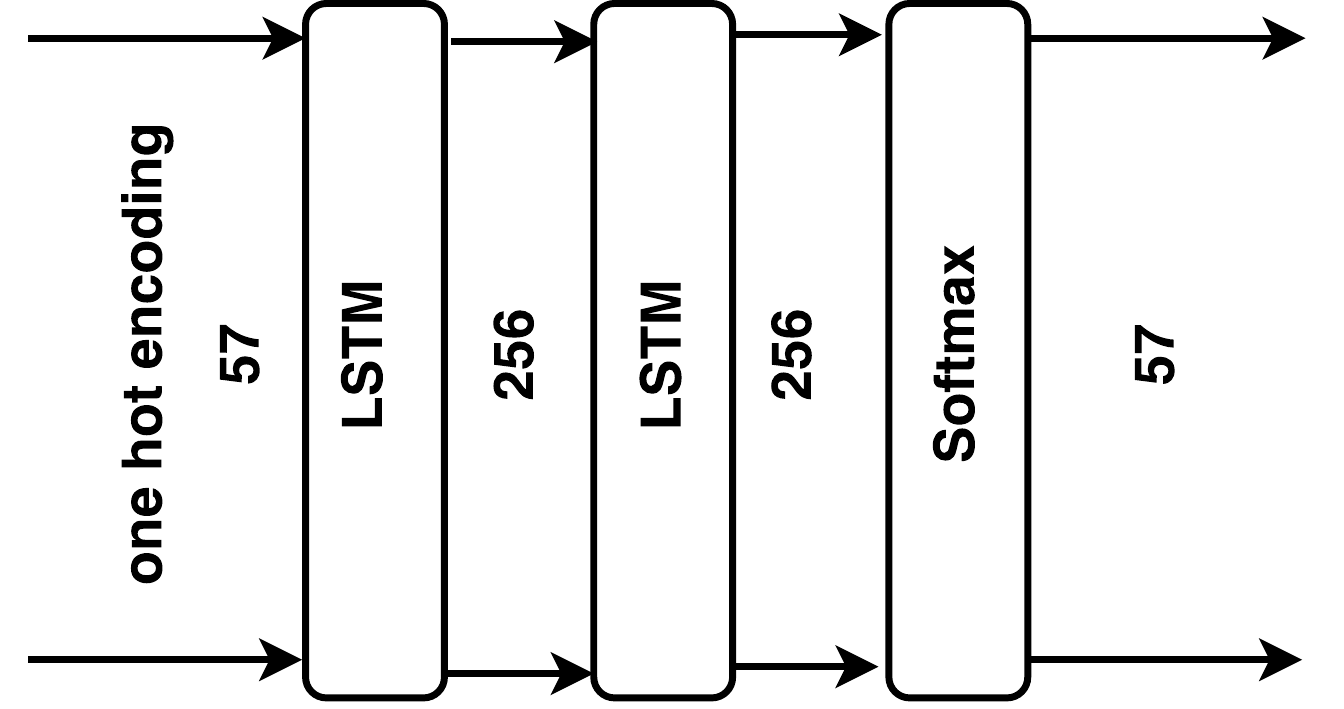}
\caption{Alphabet based neural network architecture used in DeepTingle.}
\label{fig:annArchitecture}
\end{figure}

We experimented with different architectures. Our initial intuition was to mimic the architecture of different Twitter bots. Twitter's limitation of 140 characters per tweet influenced the strategy used by most neural network trained bots. They tend to work on a character-by-character approach, producing the next character based on previous characters, not words. Similarly, our first architecture, shown in Figure \ref{fig:annArchitecture}, was inspired by this representation. The numbers in the figure represent the size of data flows between network layers. The neural network consists of 3 layers: 2 LSTM layers followed by a softmax one. A softmax layer uses softmax function to convert the neural network's output to the probability distribution of every different output class~\cite{bridle1990probabilistic}. In our case, classes are different letters. The size of input and output is 57, because that's the total number of different characters in Chuck Tingle's novels. Input is represented as one hot encoding, which represents data as a vector of size $n$, where $n-1$ values are $0$'s, and only one value is $1$, signaling the class the input belongs to. 

\begin{figure}[!t]
\centering
\includegraphics[width=\linewidth]{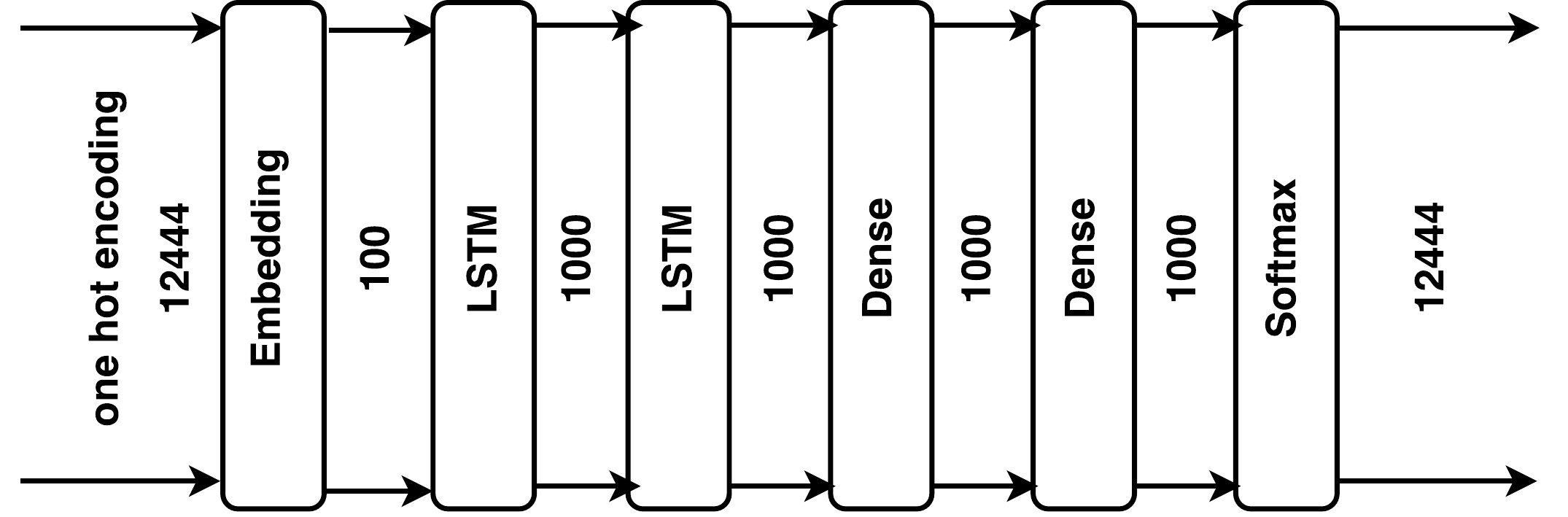}
\caption{Word-based neural network architecture used in DeepTingle.}
\label{fig:nnArchitecture}
\end{figure}

After initial testing, we opted to switch to a word representation instead of character representation. While word-based architectures repress the network's ability of creating new words, they leverage the network's sequence learning. Figure \ref{fig:nnArchitecture} shows the current architecture used in DeepTingle. The network consists of 6 layers. The first layer is an embedding one that converts an input word into its 100 dimension representation. It is followed by 2 LSTM layers of size 1000, which in turn are followed by 2 fully connected layers of same size. Finally, there is a softmax layer of size 12,444 (the total number of different words in all Tingle's books).

\subsection{Network training}

\begin{figure}
\centering
\includegraphics[width=\linewidth]{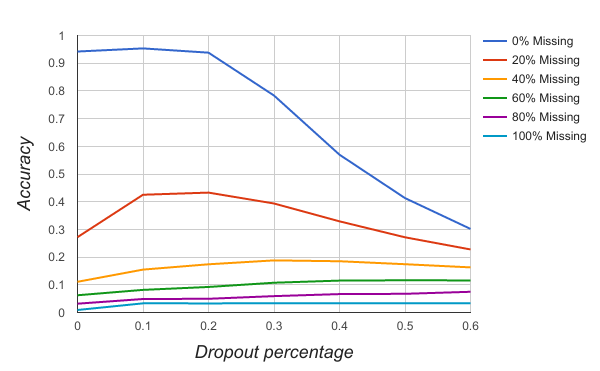}
\caption{Graph shows the effect of using dropout against noise.}
\label{fig:dropout}
\end{figure}

The network training consisted of two phases. The first one aims at training the embedding layer separately, using GloVe and all Chuck Tingle's stories in the corpus. In the second phase, we trained the remaining part of the network. Our reasoning for such approach was to speed up the learning process. Dropout is used as it increase the network accuracy against unknown input words (missing words). Figure \ref{fig:dropout} shows the effect of the dropout on the network accuracy. The graph shows using 20\% as a dropout value gives the highest accuracy without sacrificing any accuracy at 0\% missing words.

We use a recently proposed optimization technique, the Adam Optimizer~\cite{kingma2014adam}, to train the network, with a 
fixed learning rate (0.0001). This technique reaches a minimum value faster than traditional backpropagation. We experimented with various amount of time steps for the LSTM and settled for 6 time steps, for it generated sentences that were more grammatically correct and more coherent than the other experiments. Input data is designed to predict the next word based on the previous 6 words.

\subsection{Predictive Tingle}



Predictive Tingle is a writing support tool built on top of the previously mentioned network. Its goal is to provide suggestions of what next word to write, based on what the user has written so far. It does so by preprocessesing and encoding the user's input, feeding it to the network, and decoding the highest ranked outputs, which are shown as suggestions.

As the user writes, the system undergoes two phases: substitution and suggestion. Whenever a new word is written, Predictive Tingle verifies if the word appears in a Tingle-nary, a dictionary of all words from Chuck Tingle's books. If the word appears, nothing changes in this step. Otherwise, the system searches for the word in the dictionary closest to the input, using Levenshtein's string comparison ~\cite{levenshtein1966binary}. The input is then replaced with said word.

Once the substitution phase ends, the system searches for possible suggestions. It uses the last 6 written words as input for the trained network, and suggest the word with the highest output. The user can then accept or reject the suggestion. If he/she accepts, either by pressing the 'Enter' key of clicking on the suggestion button, the word is inserted in the text, and the system returns to the beginning of the suggestion phase. Otherwise, once a new word is written, the system returns to the substitution phase.

\subsection{Tingle Classics}

Tingle Classics aims to answer the question: ``what would happen if classic literature was actually written by Chuck Tingle?'' The user can select one line from a series of opening lines from famous and/or classic books (e.g. \emph{1984} by George Orwell, or \emph{Moby-dick} by Herman Melville). The system uses the line to generate a story, by repeatedly predicting the next word in a sentence. The user can also parameterize the amount of words generated, and whether to transform words that aren't in Tingle's works into words from the corpus.

\section{Results}

This section presents our results regarding the neural network training, an user study, and the two co-creativity tools developed (Predictive Tingle and Tingle Classics). A third tool, called Tingle Translator, aimed at transferring Chuck Tingle's style of writing to any given text using NN and word embeddings. Unfortunately, the embedding space for Chuck Tingle's novels is too small in comparison to the word embedding trained from Wikipedia articles. This led to a failed attempt to have a meaningful relation between both embeddings. Using a neural network to bridge this gap wasn't a success, and as such Tingle Translator will not be discussed further in this work, remaining a possibility for future work.

\subsection{Network Training}
DeepTingle trained for 2,500 epochs using the Adam Optimizer with fixed learning rate 0.0001. After 2000 epochs there was no improvement in loss. The network reached accuracy of 95\% and an error drop from 12.0 to 0.932.

\begin{algorithm}[htb]
I was walking in the streets going to my friend's house. While I was walking, I stumbled upon the chamber and then heading out into the parking lot and calling my girlfriend to confirm my status as a normal, red blooded, American heterosexual. yet, despite my best efforts, I find myself getting turned on. whoa. Kirk says with a laugh, sensing the hardening of my cock up against his back. You getting excited back there, buddy? No. I protest, defensively. It sure doesn't feel like it. The unicorn prods with a laugh. That feels like a big fucking human cock pressed up against my back. I don't say a word, completely embarrassed. You ever fucked a unicorn? Kirk asks me suddenly. I can immediately sense a change in his tone, a new direction in his unicorn mannerisms all the way down to the way the he turns his large beastly head to speak to me. No, I can't say that i have. I explain. You're the first one I've met. Kirk nods. Yep, there's not a lot of us out there, not a lot of gay one's either.
\caption{Generated story where every new word depends on the previous 6 words.}
\label{ex:differentText5}
\end{algorithm}

\begin{algorithm}[ht]
I was walking in the streets going to my friend's house. While I was walking , I stumbled upon the hustle and bustle of my surroundings. instead of my win, i begin to weave out into the air with a second moments, eventually my discomfort becomes apparent and closer to the cars. suddenly, i feel the strangely gay being of chibs suddenly, only this long i try not to stare too. where am i like? i question. but, you have a point, jonah says. when i was in there for a moment, my mind drifting almost i have ever seen in this situation; no living longer in our game. as i said this was the hunk hand, and i know this about the man in a situation so much more than i have to really right about this. i understand, that's how i want to do and handsome, love. of course, it is, i really believe that i really want. ever before, i don't know. my wife explains, the rich man explains. this was amazing, i remind him. the dinosaur takes a few steps behind the top of the stage and immediately standing up the front screen.
\caption{Generated story where every new word depends on the previous 20 words.}
\label{ex:differentText20}
\end{algorithm}

We experimented with different sizes of word sequences, from 1 word up to 20 words. Examples \ref{ex:differentText5} and \ref{ex:differentText20} show chunks of generated text in 2 sizes (6 and 20 word sequence). All experiments started with the same input, i.e. \emph{``I was walking in the streets going to my friend's house . While I was walking , I stumbled upon''}, and generated at least 200 words. It is trivial to recognize that the 6 words sequence produce more grammatically correct sentences compared to the 20 words sequence. On the other hand, 20 words sequences have higher chance to refer to something that happened before, and less chances of getting stuck in loops when compared to 6 words sequences. 

\begin{figure}
\centering
\includegraphics[width=\linewidth]{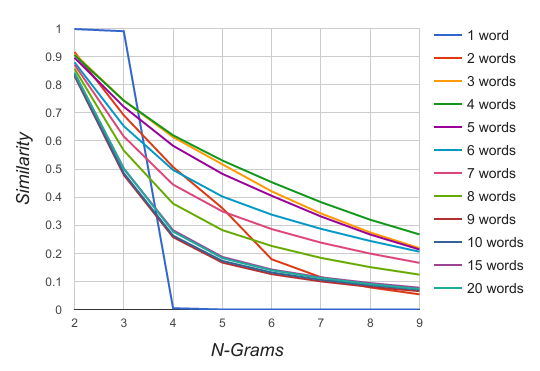}
\caption{Graph with the similarity between generated texts and the actual chuck tingle stories for all 4 sequence sizes.}
\label{fig:ngrams}
\end{figure}

To better understand the effect of increasing the sequence size, we generated a 200,000 words text, to be compared to original Chuck Tingle stories in order to evaluate how similar they are. The similarity is calculated by counting the number of identical sequence of words between the generated text and the original text. Figure \ref{fig:ngrams} shows the different N-Grams for all the sequence sizes. The 4-words sequence is the most similar to original Chuck Tingle text. Interestingly, all sizes above 8 words have the same amount of similarity. We believe this may be due to the LSTM reaching its maximum capacity at size of 9.

\begin{figure}
\centering
\includegraphics[width=\linewidth]{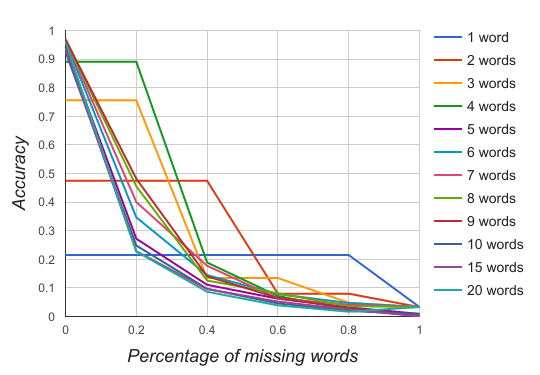}
\caption{This graph is showing the robustness of the network against missing information for all 4 sequence sizes.}
\label{fig:accuracyInput}
\end{figure}

Another experiment aimed at testing the robustness of the network, by testing the effect of unknown words on the accuracy of prediction. Figure ~\ref{fig:accuracyInput} describes the accuracy for all the sequence sizes against different percentages of missing words from the input text. It shows that the more words we have the better the results except for sizes 3 and 4. At these sizes, 20\% missing data means nothing change. We chose size 6 as it is higher than the others, and at the same time won't compromise the neural network speed.

\subsection{User Study}
\begin{table}[tb]
\centering
 \begin{tabular}{|l|c c c|} 
 \hline
  & Grammar & Coherence & Interesting \\ [0.5ex] 
 \hline
CT vs DT & $16 / 23^{*}$ & $19 / 27^{*}$ & $17 / 31$ \\ 
CT vs Markov & $29 / 31^{**}$ & $31 / 33^{**}$ & $26 / 33^{**}$ \\
DT vs Markov & $17 / 21^{**}$ & $21 / 27^{**}$ & $11 / 19$ \\
 \hline
 \end{tabular}
 \caption{Table shows the result of the user study where \textbf{CT} is Chuck Tingle's original text, \textbf{Markov} is the Markov chain generated text, and \textbf{DT} is the DeepTingle generated text. The superscript indicate the p-value from using binomial test. $^{*}$ indicated that the p-value is less than 5\%, while $^{**}$ indicates the p-value is less than 1\%.}
 \label{tab:userstudy}
\end{table}

We performed a user study to compare the generated text by DeepTingle to Chuck Tingle's original text. Additionally, we wanted to confirm if a neural network would actually have an advantage over a simpler representation, such as a Markov chain model. We trained a Markov chain on the same data set, and chose the state size to be 3 as it empirically achieved the best results without losing generalization ability.

In the user study, the user is presented with two pieces of text of equal length picked randomly from any of the 3 categories of text (Chuck Tingle's original text, DeepTingle text, and Markov chain text). The user has to answer 3 questions: ``Which text is more grammatically correct?''; ``Which text is more interesting?''; and ``Which text is more coherent?'. The user could pick one of four options: ``Left text is better'', ``Right text is better'', ``Both are the same'', or ``None''.

We collected approximately 146 different comparisons. Table ~\ref{tab:userstudy} presents the results of comparisons, excluding all choices for ``Both are the same'' or ``None of them''. The values represent the fraction of times the first text is voted over the second one. Results show that using neural networks for text prediction produce more coherent and grammatically correct text than Markov chain, but less so than the original text, which is reasonable considering the latter is written and reviewed by a human.

\subsection{Predictive Tingle}
\begin{figure}
\centering
\includegraphics[width=\linewidth]{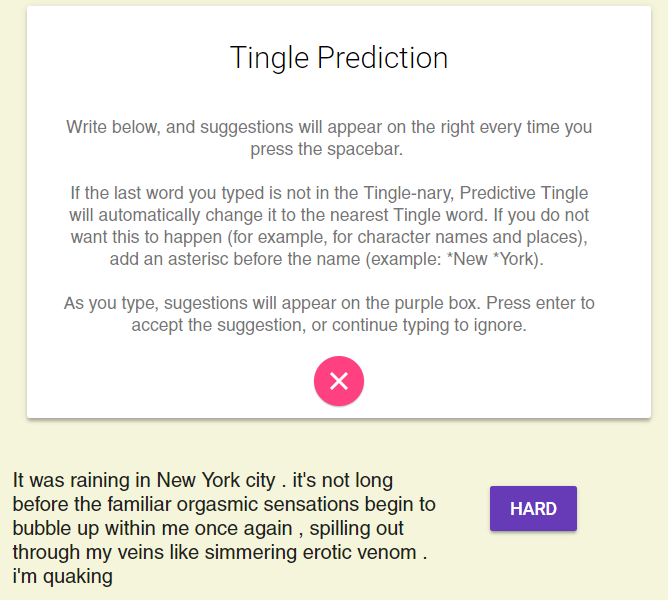}
\caption{Screenshot of Predictive Tingle. Shows the input box with an example text, and a sugestion of the next word.}
\label{fig:predictive}
\end{figure}

Figure ~\ref{fig:predictive} shows a screenshot of the system: On top we have a brief description of what Predictive Tingle is. Right below, a text field where the user can write text. To the text field's right, a purple suggestion button that is updated every time the user presses the spacebar. In this example, the user wrote ``\textit{It was raining in New York}'', and pressed enter consecutively, allowing the system to finish the input. The outcome was ``\textit{It was raining in New York city. It's not long before the familiar orgasmic sensations begin to bubble up within me once again, spilling out through my veins like simmering erotic venom.}''

\subsection{Tingle Classics}

\begin{figure}[tb]
\centering
\includegraphics[width=\linewidth]{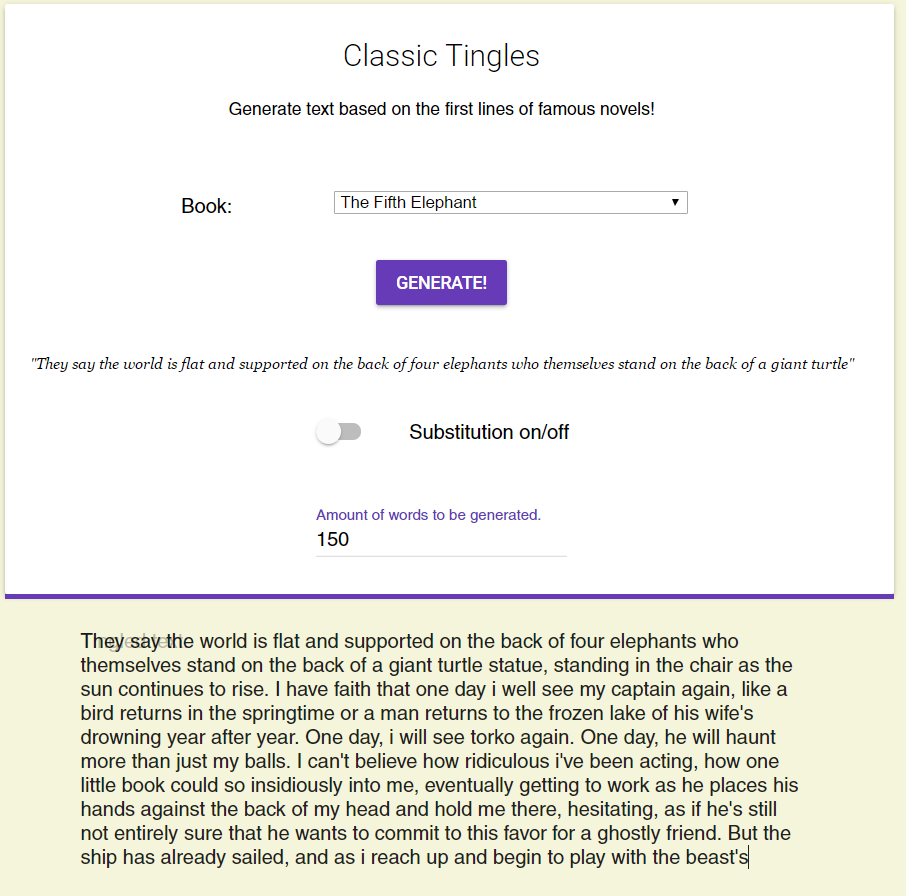}
\caption{Display of Tingle Classics, generating 150 words from the first lines in Terry Pratchet's ``The Fifth Elephant''.}
\label{fig:classic}
\end{figure}

The final part of the tools is Tingle Classics, shown in Figure ~\ref{fig:classic}. From top to bottom, the screen shows the tool's name and description, followed by a list of books, to be selected by the user. A button, "Generate!", triggers the word generation. A line, right bellow the bottom, shows the original initial line for the book selected. Two configurations options can be found in sequence: the option of toggle substitution on and off, and the amount of words to generate. Finally, the story generated is outputted at the very bottom of the page.

If substitution is selected, a preprocessing of the initial line is made, transforming every word in the original text that doesn't appear in the Tingle corpus, into a Tingle word. Thus, it guarantees that every word in the input vector appears in the Tingle corpus. If substitution is not used, words not in the Tingle corpus are skipped. For example, if the sentence is ``Hello, my name is Inigo Montoya'', and neither ``Inigo'' nor ``Montoya'' belong in the corpus, the vector would shift to embed only ``Hello, my name is'' (notice that the comma is considered a word). This may result in diverging stories, as shown in Examples ~\ref{ex:classicwithout} and ~\ref{ex:classicwith}. Both are generated from the same line (``Call me Ishmael'', from Moby-Dick, by Herman Melville), but the first doesn't use substitution, while the second does. 
 
\begin{algorithm}[ht]
Call me ishmael a simple season. The creature declares, driving the rest of his drink and then gets it in, his eyes watering tight as he thrusts into me, the massive rod filling my butthole entirely as i cry out with a yelp of pleasure. Colonel peach wastes no time now, immediately getting to work as he rams my body from behind. I grip tightly onto the bed sheets in front of me, bracing myself against the hood as slater continues to pump in and out of my butt, slowly but firmly as i tremble from his skilled touch. My legs are spread wide for him, held back as he slams into me at an ever escalating speed. Soon enough, kirk is hammering into me with everything he's got, his hips pounding loudly against the side of the boulder
\caption{150 words generated from the line ``Call me Ishmael'', without word substitution.}
\label{ex:classicwithout}
\end{algorithm}

\begin{algorithm}[ht]
Call me small new era of the night before, but somehow my vision is assaulted by sudden and graphic depictions of gay sex. I scramble to change the channel and quickly realize that every station has been somehow converted into hardcore pornography. What the fuck? I ask in startled gasp. What is this? I know that we both have a knack for running out on relationships. Portork tells me. But we also know love when we see it. A broad smile crosses my face. I see you'll also picked up my habit of inappropriate practical jokes. Portork laughs. Of course. Now get in here an fuck me, it's time for round two. Oliver explains. And i may be a country boy but i'm not stupid. I might not have the password or whatever it is that
\caption{150 words generated from the line ``Call me Ishmael'', using word substitution.}
\label{ex:classicwith}
\end{algorithm}

\section{Conclusion and Future Work}


This paper proposes a two-part system, composed of a deep neural network trained over a specific literary corpus and a writing assistance tool built on the network. Our corpus consists solely of works by renowned author Chuck Tingle. This corpus represents a large set of stories, diverse in setting and context, but similar in structure. Its controversial themes negates the ``neutral' norm of writing assistance tools currently available. We trained a six layer architecture, using GloVe embeding, LSTMs, dense and softmax layers, capable of word sequence prediction. Our system allows for users to write stories, receiving word suggestions in real time, 
and to explore the intersection of classic literature and the fantastic erotic niche that Tingle embodies.

We are excited to study how much deeper we can take DeepTingle. We intend to improve the system's architecture, in order to increase its prediction accuracy against missing words. Furthermore, a possibility is to incorporate generative techniques to evolve grammars based on Tingle's work. Additionally, we intend on improving and adding new co-creativity tools, in particular the Tingle Translator. The use case of the Tingle Translator is to take existing English text and translate it to Tingle's universe by substituting commonly used but un-Tingly words and phrases with their Tingle-equivalents. For this, we will explore different approaches to map words into embedding space, including the use of bidirectional networks and style transfer.

The central idea motivating this study and paper was to expose the norms inherent in ``neutral'' corpuses used to train AI-based assistants, such as writing assistants, and explore what happens when building a writing assistance tool trained on very non-neutral text. It is very hard to gauge the success of our undertaking through quantitative measures such as user studies. We believe that the effects of DeepTingle can best be understood by interacting with it directly, and we urge our readers to do so at their leisure.

\section{Acknowledgments}
We thank Marco Scirea, for helping us conceive ideas for this work, Philip Bontrager, for useful discussions, Scott Lee and Daniel Gopstein, for their support and enthusiasm. We gratefully acknowledge a gift of the NVidia Corporation of GPUS to the NYU Game Innovation Lab. Gabriella Barros acknowledges financial support from CAPES and the Science Without Borders program, BEX 1372713-3. Most of this paper was written by humans.






\bibliographystyle{iccc}
\bibliography{iccc}

\end{document}